\documentclass[10pt,letterpaper]{article}
\usepackage{float}
\usepackage{wacv}
\usepackage{times}
\usepackage{epsfig}
\usepackage{graphicx}
\usepackage{amsmath}
\usepackage{amssymb}
\usepackage[font=small]{caption}
\usepackage{subcaption}

\wacvfinalcopy

\begin{document}

\title{Fused DNN: A deep neural network fusion approach to fast and robust pedestrian detection}

\author{Xianzhi Du$^1$, Mostafa El-Khamy$^2$, Jungwon Lee$^2$, Larry S. Davis$^1$\\
\\
$^1$Computer Vision Laboratory, UMIACS, University of Maryland, College Park, MD 20742, USA\\
$^2$Modem Systems R$\&$D, Samsung Electronics, San Diego, CA 92121, USA\\
{\tt\small xianzhi@umiacs.umd.edu,\tt\small \{mostafa.e,jungwon2.lee\}@samsung.com,\tt\small lsd@umiacs.umd.edu}
}
\maketitle

\begin{abstract}
We propose a deep neural network fusion architecture for fast and robust pedestrian detection. The proposed network fusion architecture allows for parallel processing of multiple networks for speed. A single shot deep convolutional network is trained as a object detector to generate all possible pedestrian candidates of different sizes and occlusions. This network outputs a large variety of pedestrian candidates to cover the majority of ground-truth pedestrians while also introducing a large number of false positives. Next, multiple deep neural networks are used in parallel for further refinement of these pedestrian candidates. We introduce a soft-rejection based network fusion method to fuse the soft metrics from all networks together to generate the final confidence scores. Our method performs better than existing state-of-the-arts, especially when detecting small-size and occluded pedestrians. Furthermore, we propose a method for integrating pixel-wise semantic segmentation network into the network fusion architecture as a reinforcement to the pedestrian detector. The approach outperforms state-of-the-art methods on most protocols on Caltech Pedestrian dataset, with significant boosts on several protocols. It is also faster than all other methods.
\end{abstract}

\section{Introduction}
Pedestrian detection has applications in various areas such as video surveillance, person identification, image retrieval, and advanced driver assistance systems (ADAS). Real-time accurate detection of pedestrians is a key for adoption of such systems. A pedestrian detection algorithm aims to draw bounding boxes which describe the locations of pedestrians in an image in real-time. However, this is difficult to achieve due to the tradeoff between accuracy and speed \cite{caltech}. Whereas a low-resolution input will, in general, result in fast object detection but with poor performance, better object detection can be obtained by using a high-resolution input at the expense of processing speed. Other factors such as crowded scene, non-person occluding objects, or different appearances of pedestrians (different poses or clothing styles) also make this problem challenging.   

The general framework of pedestrian detection can be decomposed into region proposal generation, feature extraction, and pedestrian verification \cite{pedsurvey}. Classic methods commonly use sliding window based techniques for proposal generation, histograms of gradient orientation (HOG) \cite{HOG} or scale-invariant feature transform (SIFT) \cite{SIFT} as features, and support vector machine (SVM) \cite{svm} or Adaptive Boosting \cite{adaboost} as the pedestrian verification methods. Recently convolutional neural networks have been applied to pedestrian detection. Hosang et al. \cite{SCF+AlexNet} use SquaresChnFtrs \cite{tenyears} method to generate pedestrian proposals and train a AlexNet \cite{alexnet} to perform pedestrian verification. Zhang et al. \cite{rpn} use a Region Proposal Network (RPN) \cite{fasterrcnn} to compute pedestrian candidates and a cascaded Boosted Forest \cite{bf} to perform sample re-weighting to classify the candidates. Li et al. \cite{safcnn} train multiple Fast R-CNN \cite{fastrcnn} based networks to detect pedestrians with different scales and combine results from all networks to generate the final results.

\begin{figure*}
\begin{center}
   \includegraphics[width=0.8\linewidth]{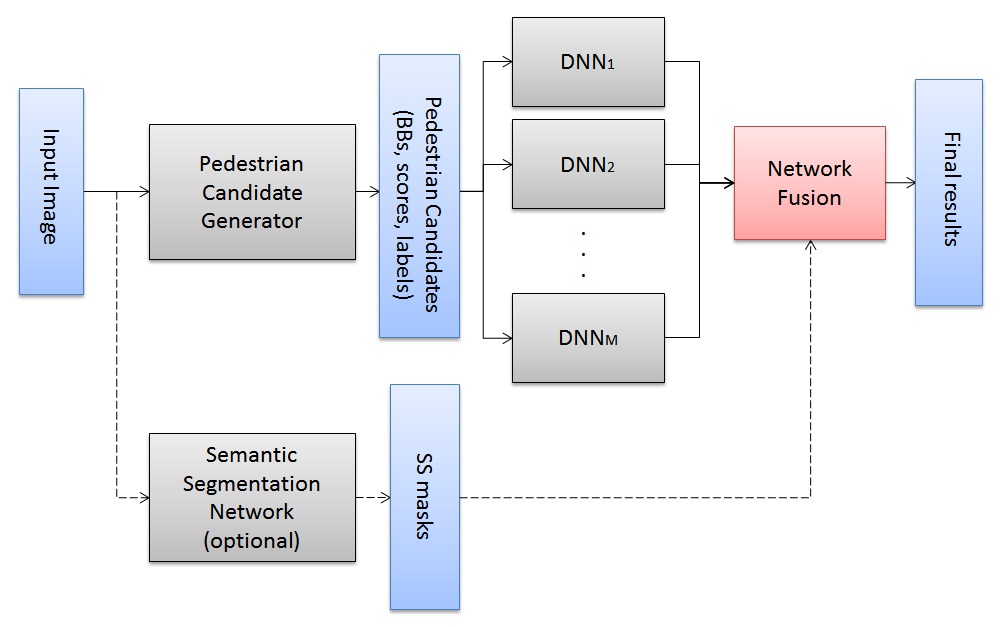}
\end{center}
   \caption{The whole pipeline of our proposed work.}
\label{fig:short}
\end{figure*}

We propose a deep neural network fusion architecture to address the pedestrian detection problem, called Fused Deep Neural Network (F-DNN). Compared to previous methods, the proposed system is faster while achieving better detection accuracy. The architecture consists of a pedestrian candidiate generator, which is obtained by training a deep convolutional neural network to have a high detection rate, albeit a large false positive rate. A novel network fusion method called soft-rejection based network fusion is proposed. It employs a classification network, consisting of multiple deep neural network classifiers, to refine the pedestrian candidates. Their soft classification probabilities are fused with the original candidates using the soft-rejection based network fusion method. A parallel semantic segmentation network, using deep dilated convolutions and context aggregation \cite{sspaper}, delivers another soft confidence vote on the pedestrian candidates, which are further fused with the candidate generator and the classification network.   

Our work is evaluated on the Caltech Pedestrian dataset \cite{caltech}. We improve log-average miss rate on the 'Reasonable' evaluation setting from 9.58\% (previous best result \cite{rpn}) to 8.65\% (8.18\% with semantic segmentation network). Meanwhile, our speed is $1.67$ times faster ($3$ times faster for 'Reasonable' test). Our numerical results show that the proposed system is accurate, robust, and efficient. 

The rest of this paper is organized as follows. Section 2 provides a detailed description of each step of our method. Section 3 discusses the experiment results and explores the effectiveness of each component of our method. Section 4 draws conclusions and discusses about potential future work.

\section{The Fused Deep Neural Network}
The proposed network architecture consists of a pedestrian candidate generator, a classification network, and a pixel-wise semantic segmentation network. The pipeline of the proposed network fusion architecture is shown in Figure 1.

For the implementation described in this paper, the candidate generator is a single shot multi-box detector (SSD) \cite{SSD}. The SSD generates a large pool of candidates with the goal of detecting all true pedestrians, resulting in a large number of false positives. Each pedestrian candidate is associated with its localization box coordinates and a confidence score. By lowering the confidence score threshold above which a detection candidate is accepted, candidates of various sizes and occlusions are generated from the primary detector. The classification network consists of multiple binary classifiers which are run in parallel. We propose a new method for network fusion called soft-rejection based network fusion (SNF). Instead of performing hard binary classification, which either accepts or rejects candidates, the confidence scores of the pedestrian candidates are boosted or discounted based on the aggregated degree of confidence in those candidates from the classifiers. We further propose a method for utilizing the context aggregation dilated convolutional network with semantic segmentation (SS) as another classifier and integrating it into our network fusion architecture. However, due to the large input size and complex network structure, the improved accuracy comes at the expense of a significant loss in speed.

\begin{figure*}
\begin{center}
   \includegraphics[width=0.8\linewidth]{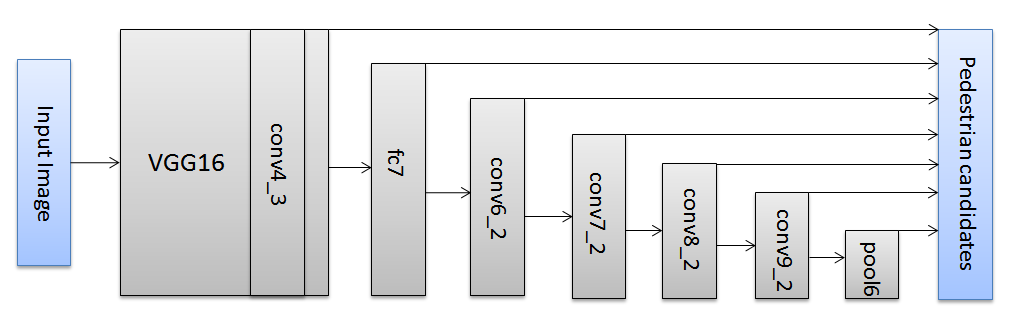}
\end{center}
   \caption{The structure of SSD. 7 output layers are used to generate pedestrian candidates in this work.}
\label{fig:short}
\end{figure*}

\subsection{Pedestrian Candidate Generator}
We use SSD to generate pedestrian candidates. The SSD is a feed-forward convolutional network which has a truncated VGG16 as the base network. In VGG16 base, pool5 is converted to $3\times 3$ with stride one, and fc6 and fc7 are converted to convolutional layers with atrous algorithm \cite{sspaper}. Additional 8 convolutional layers and a global average pooling layer are added after the base network and the size of each layer decreases progressively. Layers 'conv4\textunderscore 3', 'fc7', 'conv6\textunderscore 2', 'conv7\textunderscore 2', 'conv8\textunderscore 2', 'conv9\textunderscore 2', and 'pool6' are used as the output layers. Since 'conv4\textunderscore 3' has a much larger feature scale, an L2 normalization technique is used to scale down the feature magnitudes \cite{ParseNetLW}. After each output layer, bounding box(BB) regression and classification are performed  to generate pedestrian candidates. Figure 2 shows the structure of SSD.

For each output layer of size $m\times n\times p$, a set of default BBs at different scales and aspect ratios are placed at each location. $3\times 3\times p$ convolutional kernels are applied to each location to produce classification scores and BB location offsets with respect to the default BB locations. A default BB is labeled as positive if it has a Jaccard overlap greater than 0.5 with any ground truth BB, otherwise negative (as shown in Equation (1)).  
\begin{equation} \label{}
    label=
\begin{cases}
    1,& \text{if } \frac{A_{BB_{d}}\cap A_{BB_{g}}}{A_{BB_{d}}\cup A_{BB_{g}}}>0.5 \\ 
    0,& \text{otherwise}
\end{cases}
\end{equation}
where $A_{BB_{d}}$ and $A_{BB_{g}}$ represent the area covered by the default BB and the ground true BB, respectively. The training objective is given as Equation (2):
\begin{equation}
L=\frac{1}{N}(L_{conf}+\alpha L_{loc})
\end{equation}
where $L_{conf}$ is the softmax loss and $L_{loc}$ is the Smooth L1 localization loss \cite{fastrcnn}, $N$ is the number of positive default boxes, and $\alpha$ is a constant weight term to keep a balance between the two losses. For more details about SSD please refer to \cite{SSD}. Since SSD uses 7 output layers to generate multi-scale BB outputs, it provides a large pool of pedestrian candidates varying in scales and aspect ratios. This is very important to the following work since pedestrian candidates generated here should cover almost all the ground truth pedestrians, even though many false positives are introduced at the same time. Since SSD uses a fully convolutional framework, it is fast.

\begin{figure*}
\begin{subfigure}{.5\textwidth}
  \centering
  \includegraphics[width=.81\linewidth]{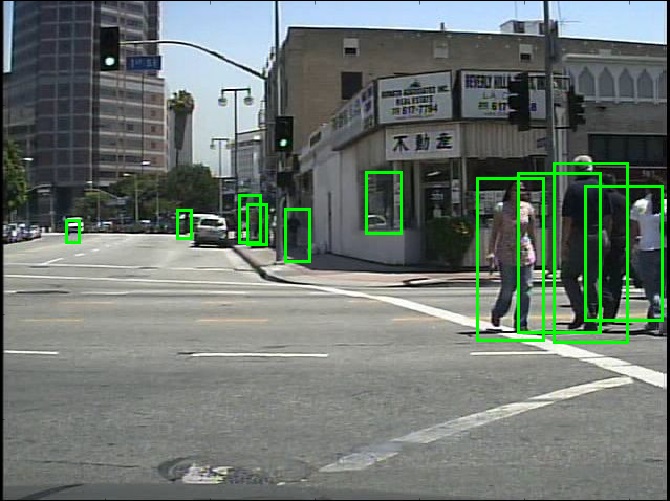}
  \label{fig:sfig1}
\end{subfigure}
\begin{subfigure}{.5\textwidth}
  \centering
  \includegraphics[width=.84\linewidth]{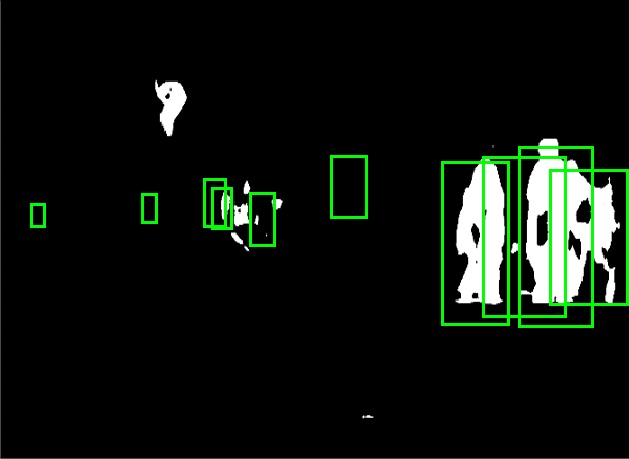}
  \label{fig:sfig2}
\end{subfigure}
\caption{Left figure shows the pedestrian candidates' BBs on an image. Right figure shows the SS mask over the BBs. We can visualize several false positives (such as windows and cars) are softly rejected by the SS mask.}
\label{fig:fig}
\end{figure*}

\subsection{Classification Network and Soft-rejection based DNN Fusion}
The classification network consists of multiple binary classification deep neural networks which are trained on the pedestrian candidates from the first stage. All candidates with confidence score greater than $0.01$ and height greater than 40 pixels are collected as the new training data for the classification network. For each candidate, we scale it to a fixed size and directly use positive/negative information collected from Equation (1) for labeling.

After training, verification methods are implemented to generate the final results. Traditional hard binary classification results in hard rejection and will eliminate a pedestrian candidate based on a single negative vote from one classification network. Instead, we introduce the SNF method which works as follows: Consider one pedestrian candidate and one classifier. If the classifier has high confidence about the candidate, we boost its original score from the candidate generator by multiplying with a confidence scaling factor greater than one. Otherwise, we decrease its score by a scaling factor less than one. We define "confident" as a classification probability of at least $a_c$. To prevent any classifier from dominating, we set $b_c$ as the lower bound for the scaling factor. Let $p_m$ be the classification probability generated by the $m_{th}$ classifier for this candidate, the scaling factor is computed as Equation (3).
\begin{equation}
a_{m}=\max(p_{m}\times \frac{1}{a_c},b_c)
\end{equation}
where $a_c$ and $b_c$ are chosen as $0.7$ and $0.1$ by cross validation. To fuse all $M$ classifiers, we multiply the candidate's original confidence score with the product of the confidence scaling factors from all classifiers in the classification network.
This can be expressed as Equation (4).
\begin{equation}
S_{FDNN}=S_{SSD}\times \prod_{m=1}^{M}a_m 
\end{equation}

The key idea behind SNF is that we don't directly accept or reject any candidates, instead we scale them with factors based on the classification probabilities. This is because a wrong elimination of a true pedestrian (e.g. as in hard-binary classification) cannot be corrected, whereas a low classification probability can be compensated for by larger classification probabilities from other classifiers.

\subsection{Pixel-wise semantic segmentation for object detection reinforcement} 
We utilize an SS network, based on deep dilated convolutions and context aggregation \cite{sspaper}, as a parallel classification network. The SS network is trained on the Cityscapes dataset for driving scene segmentation \cite{cityscapes}. To perform dense prediction, the SS network consists of a fully convolutional VGG16 network, adapted with dilated convolutions as the front end prediction module, whose output is fed to a multi-scale context aggregation module, consisting of a fully convolutional network whose convolutional layers have increasing dilation factors. 

An input image is scaled and directly processed by the SS network, which produces a binary mask with one color showing the activated pixels for the pedestrian class, and the other color showing the background. We consider both the `person' and `rider' categories in Cityscapes dataset as pedestrians, and the remaining classes as background. The SS mask is intersected with all detected BBs from the SSD. We propose a method to fuse the SS mask and the original pedestrian candidates. The degree to which each candidate's BB overlaps with the pedestrian category in the SS activation mask gives a measure of the confidence of the SS network in the candidate generator's results. We use the following strategy to fuse the results: If the pedestrian pixels occupy at least $20\%$ of the candidate BB area, we accept the candidate and keep its score unaltered; Otherwise, we apply SNF to scale the original confidence scores. This is summarized in Equation (5):
\begin{equation} \label{}
    S_{all}=
\begin{cases}
    S_{FDNN},& \text{if } \frac{A_m}{A_b}>0.2 \\ 
    S_{FDNN}\times \max(\frac{A_m}{A_b}\times a_{ss}, b_{ss}), & \text{otherwise}
\end{cases}
\end{equation}
where $A_{b}$ represents the area of the BB, $A_m$ represents the area within $A_b$ covered by semantic segmentation mask, $a_{ss}$, and $b_{ss}$ are chosen as $4$ and $0.35$ by cross validation. As we don't have pixel-level labels for pedestrian detection datasets, we directly implement the SS model \cite{sspaper} trained on the Cityscape dataset \cite{cityscapes}. Figure 3 shows an example of this method and how we fuse it into the existing model.

SNF with an SS network is slightly different from SNF with a classification network. The reason is that the SS network can generate new detections which have not been produced by the candidate generator, which is not the case for the classification network. To address this, the proposed SNF method eliminates new detections from the SS network. The idea is illustrated in Figure 4.

\begin{figure}
\begin{center}
   \includegraphics[width=0.5\linewidth]{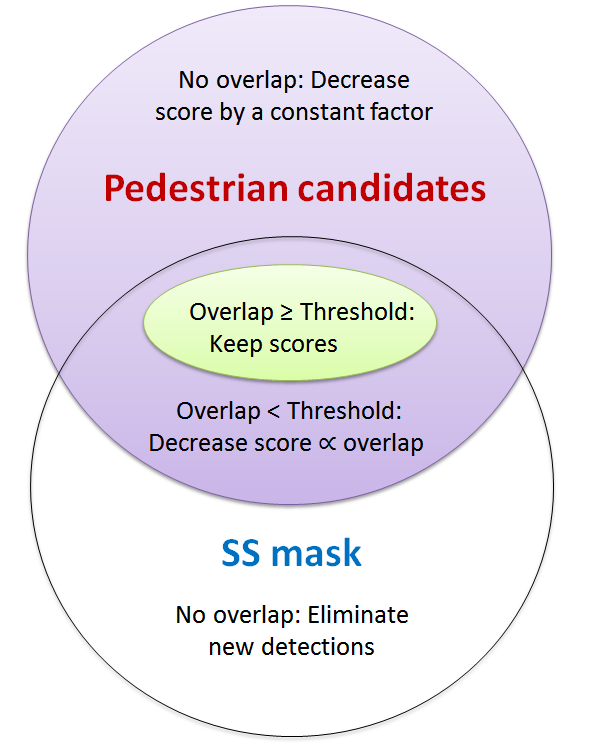}
\end{center}
   \caption{Implementing semantic segmentation as a reinforcement to pedestrian detection problem.}
\label{fig:short}
\end{figure}

\section{Experiments and result analysis}

\subsection{Data and evaluation settings}
We evaluate the proposed method on the most popular pedestrian detection dataset: Caltech Pedestrian dataset. The Caltech Pedestrian dataset contains 11 sets (S0-S10), where each set consists of 6 to 13 one-minute long videos collected from a vehicle driving through an urban environment. There are about 250,000 frames with about 350,000 annotated BBs and 2300 unique pedestrians. Each BB is assigned with one of the three labels: 'Person', 'People' (large group of individuals), and 'Person?' (unclear identifications). 
The original frame size is  $480\times 640$. The log-average miss rate (L-AMR) is used as the performance evaluation metric \cite{caltech}. L-AMR is computed evenly spaced in log-space in the range $10^{-2}$ to $10^0$ by averaging miss rate at the rate of nine false positives per image (FPPI) \cite{caltech}. There are multiple evaluation settings defined based on the height and visible part of the BBs. The most popular settings are listed in Table 1. 

\begin{table}[h!]
\begin{center}
\begin{tabular}{|l|l|l|}
\hline
Setting & Description\\
\hline\hline
Reasonable & 50+ pixels. Occ. none or partial\\
All & 20+ pixels. Occ. none, partial, or heavy\\
Far & 30- pixels\\
Medium & 30-80 pixels\\
Near & 80+ pixels\\
Occ. none & 0\% occluded\\
Occ. partial & 1-35\% occluded\\
Occ. heavy & 35-80\% occluded\\
\hline
\end{tabular}
\end{center}
\caption{Evaluation settings for Caltech Pedestrian dataset.}
\end{table}

\begin{table*}[h!]
\begin{center}
\begin{tabular}{|l|l|l|l|l|l|l|l|l|}
\hline
Method & Reasonable & All & Far & Medium & Near & Occ. none & Occ. partial & Occ. heavy\\
\hline\hline
SCF+AlexNet \cite{SCF+AlexNet} & 23.32\% & 70.33\% & 100\% & 62.34\% &  10.16\% & 19.99\% & 48.47\% & 74.65\%\\
SAF R-CNN \cite{safcnn} & 9.68\% & 62.6\% & 100\% & 51.8\% & \textbf{0\%} & 7.7\% & 24.8\% & 64.3\%\\
MS-CNN \cite{mscnn} & 9.95\% & 60.95\% & 97.23\% &  49.13\% & 2.60\% & 8.15\% & 19.24\% & 59.94\%\\
DeepParts \cite{DeepParts2015} & 11.89\% & 64.78\% & 100\% & 56.42\% & 4.78\% & 10.64\% & 19.93\% & 60.42\%\\
CompACT-Deep \cite{CompACT2015} & 11.75\% & 64.44\% & 100\% & 53.23\% & 3.99\% & 9.63\% & 25.14\% & 65.78\%\\
RPN+BF \cite{rpn} & 9.58\% & 64.66\% & 100\% & 53.93\% & 2.26\% & 7.68\% & 24.23\% & 69.91\%\\
F-DNN (Ours) & \textbf{8.65\%} & \textbf{50.55\%} & \textbf{77.37\%} & \textbf{33.27\%} & 2.96\% & \textbf{7.10\%} & \textbf{15.41\%} & \textbf{55.13\%}\\
F-DNN+SS (Ours) & \textbf{8.18\%} & \textbf{50.29\%} & \textbf{77.47\%} & \textbf{33.15\%} & 2.82\% & \textbf{6.74\%} & \textbf{15.11\%} & \textbf{53.76\%} \\
\hline
\end{tabular}
\end{center}
\caption{Detailed breakdown performance comparisons of our models and other state-of-the-art models on the 8 evaluation settings. All numbers are reported in L-AMR.}
\end{table*}

\subsection{Training details and results}
To train the SSD candidate generator, all images which contain at least one annotated pedestrian from Caltech training set, ETH dataset \cite{eth}, and TudBrussels dataset \cite{Tudbrussel} are used. By using both original and flipped images, it provides around 68,000 images. Among all annotations, only 'Person' and 'People' categories are included. We further classify 'Person' into 'Person\textunderscore full' and 'Person\textunderscore occluded'. This results in 109,000 pedestrians in 'Person\textunderscore full', 60,000 pedestrians in 'Person\textunderscore occluded', and 35,000 pedestrians in 'People'. All the images are in their original size $480\times 640$. To set up the SSD model, we place 7 default BBs with aspect ratios [0.1, 0.2, 0.41a, 0.41b, 0.8, 1.6, 3.0] on top of each location of all output feature maps. All default BBs except 0.41b have relative heights [0.05, 0.1, 0.24, 0.38, 0.52, 0.66, 0.80] for the 7 output layers. The heights for 0.41b are [0.1, 0.24, 0.38, 0.52, 0.66, 0.80, 0.94]. Since 0.41 is the average aspect ratio for all annotated pedestrians, we use two default BBs with slightly different heights. The aspect ratio '1.6' and '3.0' are designed for 'People'. By doing so, we can generate a rich pool of candidates so as not to lose any ground truth pedestrians. We then fine-tune our SSD model from the Microsoft COCO \cite{coco} pre-trained SSD model for 40k iterations using stochastic gradient descent (SGD), with a learning rate of $10^{-5}$. All the layers after each output layer are randomly initialized and trained from scratch.   

To train the classification network, we use all pedestrian cadidates generated by SSD as well as all ground truth BBs. All the training samples are horizontally flipped with probability $0.5$. This results in $69,000$ positive samples and $163,000$ negative samples, with a ratio $1:2.4$. As the high similarity between consecutive frames sometimes leads to identical training samples, we scale all samples to a fixed size of $250\times 250$, and then randomly crop a $224\times 224$ patch for data augmentation (center cropping for testing). We then fine-tune two models in parallel: ResNet-50 \cite{res50} and GoogleNet \cite{googlenet}, from their ImageNet pre-trained models. All models are trained using SGD with a learning rate of $10^{-4}$. 

For the SS network, since there is a lack of well-labeled SS dataset for pedestrian detection, we directly implement the network trained on the Cityscapes dataset with image size $1024\times 2048$. To preserve the aspect ratio, we scale all images to height $1024$, then pad black pixels on both sides. 

All the above models are built on Caffe deep learning framework \cite{caffe}.

\textbf{Evaluation on Caltech Pedestrian}: The detailed breakdown performance of our two models (without and with semantic segmentation) on this dataset is shown in Table 2. We compare with all the state-of-the-art methods reported on Caltech Pedestrian website. We can see that both of our models significantly outperform others on almost all evaluation settings. On the 'Reasonable' setting, our best model achieves $8.18\%$ L-AMR, which has a $14.6\%$ relative improvement from the previous best result $9.58\%$ by RPN+BF. On the 'All' evaluation setting, we achieve $50.29\%$, a relative improvement of $17.5\%$ from $60.95\%$ by MS-CNN \cite{mscnn}. The L-AMR VS. FPPI plots for the 'Reasonable' and 'All' evaluation settings are shown in Figure 5 and Figure 6. F-DNN refers to fusing the SSD with the classification network, whereas F-DNN+SS refers to fusing the SSD with both the classification network and the SS network. Results from VJ \cite{Viola} and HOG \cite{HOG} are plotted as the baselines on this dataset.

\subsection{Result analysis}

\subsubsection{Effectiveness of classification network}
We explore how effective the classification network refines the original confidence scores of the pedestrian candidates. As many false positives are introduced from SSD, the main goal of the classification network is to decrease the scores of the false positives. By using ResNet-50 with classification probability $0.7$ as the confidence scaling threshold, $96.7\%$ scores of the false positives are decreased on the Caltech Pedestrian testing set. This improves the performance on the 'Reasonable' setting from $13.07\%$ to $8.98\%$. A similar result is obtained by GoogleNet-$97.1\%$ scores of the false positives are decreased, which improves performance from $13.07\%$ to $9.41\%$. As the two classifiers do not decrease scores of the same set of false positives, by fusing their results with SNF, almost all false positives are covered, and the L-AMR is further improved to $8.65\%$. Concerning limits to performance, if we were able to train an oracle classifier with classification accuracy $100\%$, the L-AMR would be improved to only $4\%$. This is shown in Table 3.

\begin{table}[h!]
\begin{center}
\begin{tabular}{|l|l|}
\hline
Method & Reasonable\\
\hline\hline
SSD & 13.06\%\\
SSD+GoogleNet & 9.41\%\\
SSD+ResNet-50 & 8.97\%\\
SSD+GoogleNet+ResNet-50 (F-DNN)& 8.65\%\\
SSD+Oracle classifier & 4\%\\
\hline
\end{tabular}
\end{center}
\caption{Effectiveness of the classification network.}
\end{table}

\subsubsection{Soft rejection versus hard rejection}
SNF plays an important role in our system. Hard rejection is defined as eliminating any candidate which is classified as a false positive by any of the classifiers. The performance of hard-rejection based fusion depends on the performance of all classifiers. A comparison between the two methods is shown in Table 4. We also compare against the case when the SSD is fused with the SS network only, labeled as 'SSD+SS'. For the classification network, a $0.5$ classification probability threshold is used for hard rejection, while for the SS network, an overlap ratio of $5\%$ is used. 
We can see that hard rejection hurts performance significantly, especially for the classification network. All numbers are reported in L-AMR on the 'Reasonable' evaluation setting.

\begin{table}[h!]
\begin{center}
\begin{tabular}{|l|l|l|}
\hline
Method & Hard rejection & Soft rejection\\
\hline\hline
SSD+SS & 13.4\% & 11.57\%\\
F-DNN & 20.67\% & 8.65\%\\
F-DNN+SS & 22.33\%\% & 8.18\%\\
\hline
\end{tabular}
\end{center}
\caption{Performance comparisons on Caltech 'Reasonable' setting between soft rejection and hard rejection. The original L-AMR of SSD alone is $13.06\%$}
\end{table}

\subsubsection{Robustness on challenging scenarios}
The proposed method performs much better than all other methods on challenging scenarios such as the small pedestrian scenario, the occluded pedestrian scenario, crowded scenes, and the blurred input image. Figure 7 visualizes the results of the ground truth annotation, our method, and RPN-BF (previous state-of-the-art method). The four rows represent the four challenging scenarios and the four columns represent the BBs from the ground true annotations, the pedestrian candidates generated by SSD alone, our final detection results, and the results from RPN-BF method. By comparing the third column with the second column, we can see that the classification network and the SS network are able to filter out most of the false positives introduced by the SSD detector. By comparing the third column with the last column, we can see our method is more robust and accurate on the challenging scenarios than RPN-BF method.

\begin{figure}
\begin{center}
   \includegraphics[width=0.75\linewidth]{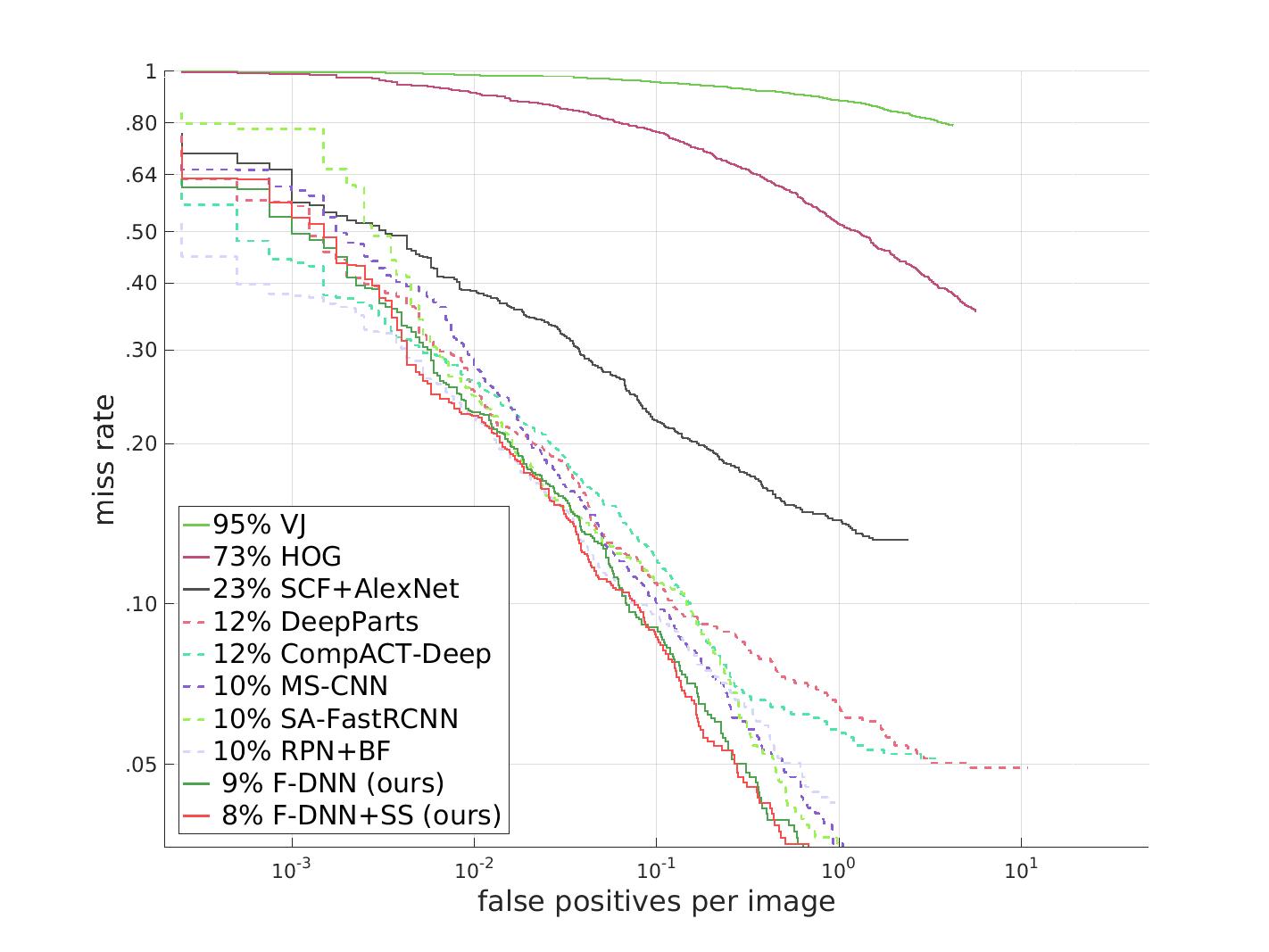}
\end{center}
   \caption{L-AMR VS. FPPI plot on 'Reasonable' evaluation setting.}
\label{fig:short}
\end{figure}

\begin{figure}
\begin{center}
   \includegraphics[width=0.75\linewidth]{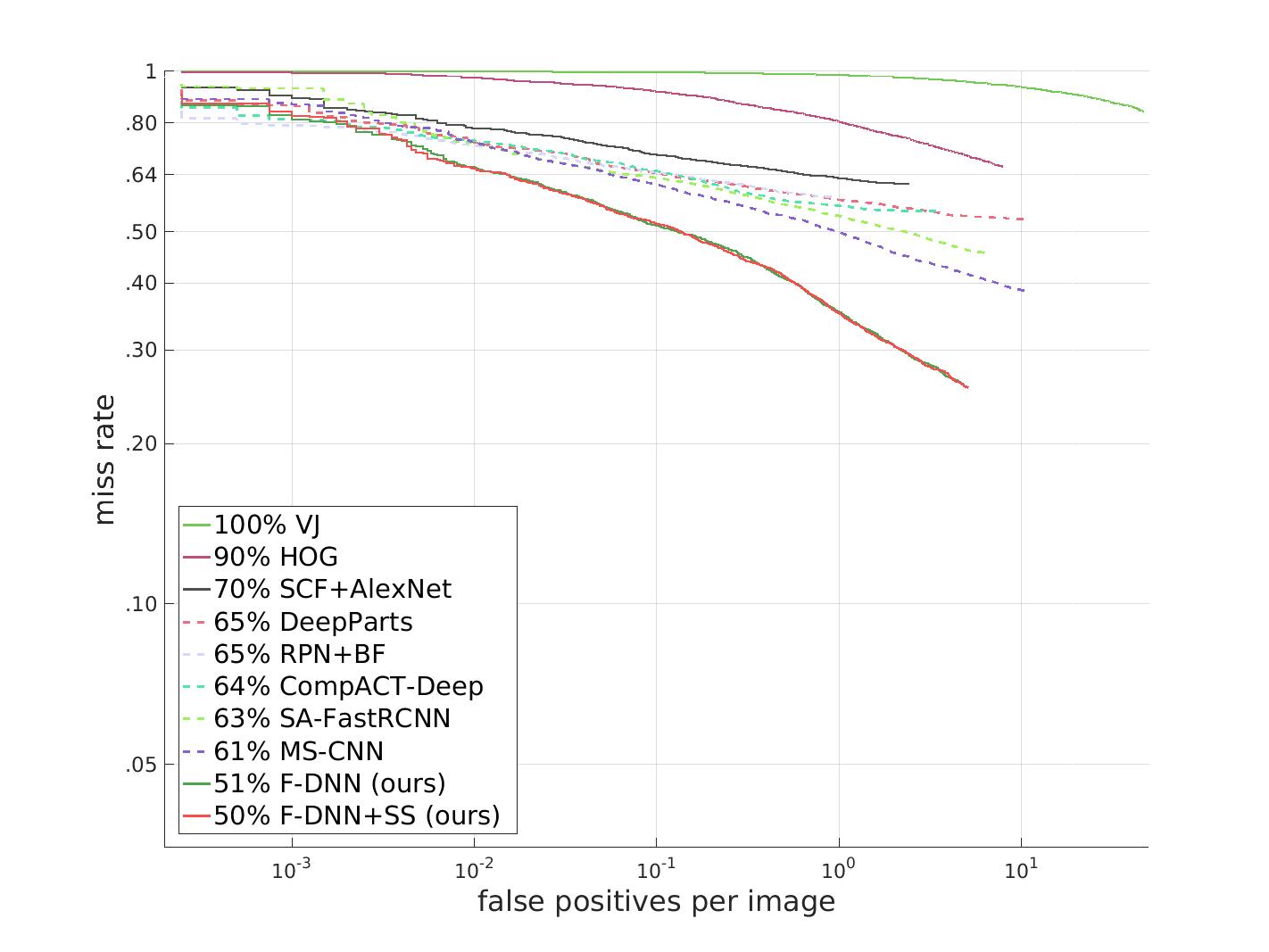}
\end{center}
   \caption{L-AMR VS. FPPI plot on 'All' evaluation setting.}
\label{fig:short}
\end{figure}

\begin{figure*}
\begin{center}
   \includegraphics[width=1\linewidth]{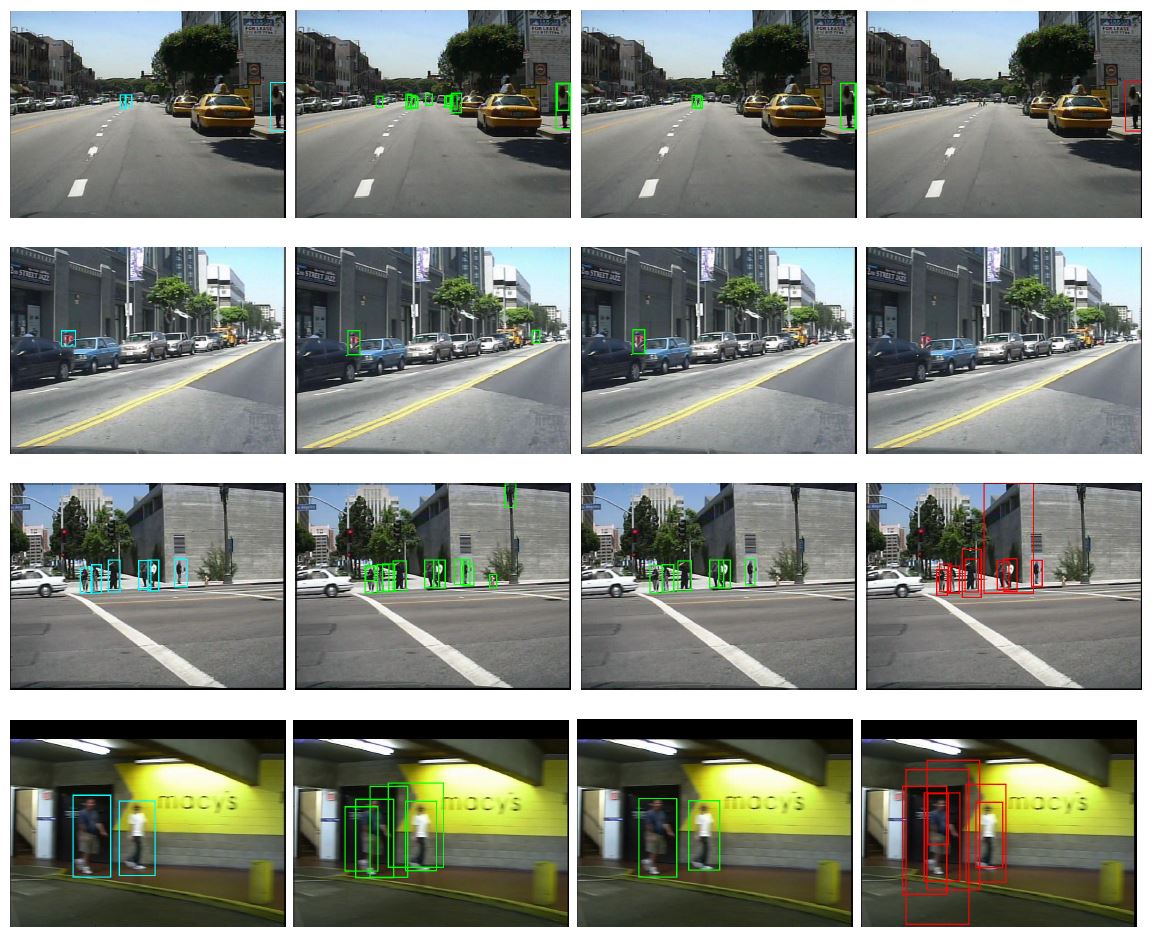}
\end{center}
\caption{Detection comparisons on four challenging pedestrian detection scenarios. The four rows represent the small pedestrian scenario, the occluded pedestrian scenario, the crowed scenes, and the blurred input image. The four columns represent the ground true annotations, the pedestrian candidates generated by SSD alone, our final detection results, and the results from the RPN-BF method.}
\label{fig:fig}
\end{figure*}

\subsubsection{Speed analysis}
There are 4 networks in the proposed work. As the SSD uses a fully convolutional framework, its processing time is $0.06$s per image. For the speed of the classification network, we perform two tests: The first test runs on all pedestrian candidates; The second test runs only on candidates above 40 pixels in height. The second test is targeting only the 'Reasonable' evaluation setting. Using parallel processing, the speed for the classification network equals its slowest classifier, which is $0.24$s and $0.1$s per image for the two tests. The overall processing time of F-DNN is $0.3$s and $0.16$s per image, which is $1.67$ to $3$ times faster than other methods. Note that the speed for GoogleNet on 'Reasonable' test is only $0.05$s per image. If we only fuse the SSD with GoogleNet, we can still achieve the state-of-the-art performance while being $0.11$s per image. We also test fusing SSD with SqueezeNet \cite{SqueezeNet} only to achieve $0.09$s per image, which is above 10 frames per second (with a L-AMR around $10.8\%$). As the SS network uses $1024\times2048$ size input with a more complex network structure, it's processing time is $2.48$s per image. As it can be processed in parallel with the whole F-DNN pipeline, the overall processing time will be limited to $2.48$s per image. We use one NVIDIA TITIAN X GPU for all the speed tests. All the classifiers in the classification network are processed in parallel on one single GPU. Table 5 compares the processing speed of our methods and the other methods.

\begin{table}
\begin{center}
\begin{tabular}{|l|l|}
\hline
Method & Speed on TITAN X \\
&(seconds per image)\\
\hline\hline
CompACT-Deep & 0.5\\
SAF R-CNN & 0.59\\
RPN+BF & 0.5\\
F-DNN & \textbf{0.3}\\
F-DNN (Reasonable) & \textbf{0.16} \\
SSD+GoogleNet (Reasonable) & \textbf{0.11} \\
SSD+SqueezeNet (Reasonable) & \textbf{0.09} \\
F-DNN+SS & 2.48\\
\hline
\end{tabular}
\end{center}
\caption{A comparison of speed among the state-of-the-art models.}
\end{table}

\section{Conclusion and Discussion}
We presented an effective and robust pedestrian detection system based on the fusion of multiple well-trained DNNs. We trained an SSD for rich pedestrian candidates generation. Then we designed a soft-rejection based network fusion approach to fuse binary classifiers based on ResNet-50 and GoogleNet to refine the candidates. We also proposed a method of using semantic segmentation to improve the pedestrian detection.

Experiments with different settings on the most popular dataset showed our method works well on pedestrians of different scales and occlusions. We outperformed all previous models, while also being faster than them.  

For future work, we want to explore the fusion of more classifiers into our existing model to make it even stronger. To train the classifiers, instead of using binary labeling strategy, we want to explore the label smoothing method introduced in \cite{inceptionV3}. Moreover, since we showed in this work that semantic segmentation can help solve the object detection problem, we would also like to explore how an object detection algorithm can help solve the instance-aware semantic labeling problem \cite{uhrig2016pixel}. 

\vspace{2\baselineskip}
\textbf{Acknowledgment} Thanks to Arvind Yedla and Marcel Nassar for the helpful discussions.
\vspace{1.5\baselineskip}

{
\bibliographystyle{IEEE}
\bibliography{FusedDNN}

\begin{thebibliography}{10}\itemsep=-1pt

\bibitem{tenyears}
R.~Benenson, M.~Omran, J.~H. Hosang, and B.~Schiele.
\newblock Ten years of pedestrian detection, what have we learned?
\newblock {\em CoRR}, abs/1411.4304, 2014.

\bibitem{svm}
C.~J. Burges.
\newblock A tutorial on support vector machines for pattern recognition.
\newblock volume~2, pages 121--167, January 1998.

\bibitem{mscnn}
Z.~Cai, Q.~Fan, R.~Feris, and N.~Vasconcelos.
\newblock A unified multi-scale deep convolutional neural network for fast
  object detection.
\newblock In {\em ECCV}, 2016.

\bibitem{CompACT2015}
Z.~Cai, M.~Saberian, and N.~Vasconcelos.
\newblock Learning complexity-aware cascades for deep pedestrian detection.
\newblock In {\em ICCV}, 2015.

\bibitem{cityscapes}
M.~Cordts, M.~Omran, S.~Ramos, T.~Rehfeld, M.~Enzweiler, R.~Benenson,
  U.~Franke, S.~Roth, and B.~Schiele.
\newblock The {Cityscapes} dataset for semantic urban scene understanding.
\newblock In {\em Proc. of the IEEE Conference on Computer Vision and Pattern
  Recognition (CVPR)}, 2016.

\bibitem{HOG}
N.~Dalal and B.~Triggs.
\newblock Histograms of oriented gradients for human detection.
\newblock In {\em In CVPR}, pages 886--893, 2005.

\bibitem{bf}
P.~Dollar.
\newblock Quickly boosting decision trees - pruning underachieving features
  early.
\newblock In {\em ICML}. International Conference on Machine Learning, June
  2013.

\bibitem{caltech}
P.~Doll\'ar, C.~Wojek, B.~Schiele, and P.~Perona.
\newblock Pedestrian detection: An evaluation of the state of the art.
\newblock {\em PAMI}, 34, 2012.

\bibitem{pedsurvey}
M.~Enzweiler and D.~M. Gavrila.
\newblock Monocular pedestrian detection: Survey and experiments.
\newblock {\em IEEE Trans. Pattern Anal. Mach. Intell.}, 31(12):2179--2195,
  Dec. 2009.

\bibitem{eth}
A.~Ess, B.~Leibe, K.~Schindler, , and L.~van Gool.
\newblock A mobile vision system for robust multi-person tracking.
\newblock In {\em IEEE Conference on Computer Vision and Pattern Recognition
  (CVPR'08)}. IEEE Press, June 2008.

\bibitem{adaboost}
Y.~Freund and R.~E. Schapire.
\newblock A decision-theoretic generalization of on-line learning and an
  application to boosting.
\newblock {\em J. Comput. Syst. Sci.}, 55(1):119--139, Aug. 1997.

\bibitem{fastrcnn}
R.~Girshick.
\newblock Fast {R-CNN}.
\newblock In {\em International Conference on Computer Vision ({ICCV})}, 2015.

\bibitem{res50}
K.~He, X.~Zhang, S.~Ren, and J.~Sun.
\newblock Deep residual learning for image recognition.
\newblock {\em arXiv preprint arXiv:1512.03385}, 2015.

\bibitem{SCF+AlexNet}
J.~H. Hosang, M.~Omran, R.~Benenson, and B.~Schiele.
\newblock Taking a deeper look at pedestrians.
\newblock {\em CoRR}, abs/1501.05790, 2015.

\bibitem{SqueezeNet}
F.~N. Iandola, M.~W. Moskewicz, K.~Ashraf, S.~Han, W.~J. Dally, and K.~Keutzer.
\newblock Squeezenet: Alexnet-level accuracy with 50x fewer parameters and
  $<$1mb model size.
\newblock {\em arXiv:1602.07360}, 2016.

\bibitem{caffe}
Y.~Jia, E.~Shelhamer, J.~Donahue, S.~Karayev, J.~Long, R.~Girshick,
  S.~Guadarrama, and T.~Darrell.
\newblock Caffe: Convolutional architecture for fast feature embedding.
\newblock {\em arXiv preprint arXiv:1408.5093}, 2014.

\bibitem{alexnet}
A.~Krizhevsky, I.~Sutskever, and G.~E. Hinton.
\newblock Imagenet classification with deep convolutional neural networks.
\newblock In F.~Pereira, C.~J.~C. Burges, L.~Bottou, and K.~Q. Weinberger,
  editors, {\em Advances in Neural Information Processing Systems 25}, pages
  1097--1105. Curran Associates, Inc., 2012.

\bibitem{safcnn}
J.~Li, X.~Liang, S.~Shen, T.~Xu, and S.~Yan.
\newblock Scale-aware fast {R-CNN} for pedestrian detection.
\newblock {\em CoRR}, abs/1510.08160, 2015.

\bibitem{coco}
T.~Lin, M.~Maire, S.~J. Belongie, L.~D. Bourdev, R.~B. Girshick, J.~Hays,
  P.~Perona, D.~Ramanan, P.~Doll{\'{a}}r, and C.~L. Zitnick.
\newblock Microsoft {COCO:} common objects in context.
\newblock {\em CoRR}, abs/1405.0312, 2014.

\bibitem{SSD}
W.~Liu, D.~Anguelov, D.~Erhan, C.~Szegedy, S.~Reed, C.-Y. Fu, and A.~C. Berg.
\newblock {SSD}: Single shot multibox detector.
\newblock {\em arXiv:1512.02325}, 2015.

\bibitem{ParseNetLW}
W.~Liu, A.~Rabinovich, and A.~C. Berg.
\newblock Parsenet: Looking wider to see better.
\newblock {\em CoRR}, abs/1506.04579, 2015.

\bibitem{SIFT}
D.~G. Lowe.
\newblock Distinctive image features from scale-invariant keypoints.
\newblock {\em Int. J. Comput. Vision}, 60(2):91--110, Nov. 2004.

\bibitem{fasterrcnn}
S.~Ren, K.~He, R.~Girshick, and J.~Sun.
\newblock Faster {R-CNN}: Towards real-time object detection with region
  proposal networks.
\newblock In {\em Advances in Neural Information Processing Systems ({NIPS})},
  2015.

\bibitem{googlenet}
C.~Szegedy, W.~Liu, Y.~Jia, P.~Sermanet, S.~E. Reed, D.~Anguelov, D.~Erhan,
  V.~Vanhoucke, and A.~Rabinovich.
\newblock Going deeper with convolutions.
\newblock {\em CoRR}, abs/1409.4842, 2014.

\bibitem{inceptionV3}
C.~Szegedy, V.~Vanhoucke, S.~Ioffe, J.~Shlens, and Z.~Wojna.
\newblock Rethinking the inception architecture for computer vision.
\newblock {\em CoRR}, abs/1512.00567, 2015.

\bibitem{DeepParts2015}
Y.~Tian, P.~Luo, X.~Wang, and X.~Tang.
\newblock Deep learning strong parts for pedestrian detection.
\newblock In {\em ICCV}, 2015.

\bibitem{uhrig2016pixel}
J.~Uhrig, M.~Cordts, U.~Franke, and T.~Brox.
\newblock Pixel-level encoding and depth layering for instance-level semantic
  labeling.
\newblock {\em arXiv preprint arXiv:1604.05096}, 2016.

\bibitem{Viola}
P.~Viola and M.~J. Jones.
\newblock Robust real-time face detection.
\newblock {\em Int. J. Comput. Vision}, 57(2):137--154, May 2004.

\bibitem{Tudbrussel}
C.~Wojek, S.~Walk, and B.~Schiele.
\newblock Multi-cue onboard pedestrian detection.
\newblock In {\em IEEE Conference on Computer Vision and Pattern Recognition
  (CVPR)}, June 2009.

\bibitem{sspaper}
F.~Yu and V.~Koltun.
\newblock Multi-scale context aggregation by dilated convolutions.
\newblock {\em To appear in ICLR 2016}, 2016.

\bibitem{rpn}
L.~Zhang, L.~Lin, X.~Liang, and K.~He.
\newblock Is faster {R-CNN} doing well for pedestrian detection?
\newblock {\em To appear in ECCV 2016}, 2016.

\end{thebibliography}
}

\end{document}